\def\year{2020}\relax
\documentclass[letterpaper]{article} 
\usepackage{aaai20}  
\usepackage{times}  
\usepackage{helvet} 
\usepackage{courier}  
\usepackage[hyphens]{url}  
\usepackage{graphicx} 
\usepackage{latexsym}
\usepackage{multirow}
\usepackage{booktabs}
\usepackage{amsmath}
\usepackage{amsmath,bm}
\usepackage{url}
\usepackage{hyperref}
\usepackage{url}
\usepackage{colortbl}
\usepackage[htt]{hyphenat}
\usepackage{mathrsfs}
\usepackage{booktabs}       
\usepackage{amsfonts}       
\usepackage{nicefrac}       
\usepackage{microtype}      
\usepackage{verbatim}
\usepackage{amssymb}
\usepackage{pifont}
\usepackage{algorithm,algorithmicx,algpseudocode}
\usepackage{lipsum}
\usepackage{graphics}
\usepackage{subfigure}
\usepackage{xcolor}
\usepackage{textcomp}
\let\OldStatex\Statex
\renewcommand{\Statex}[1][3]{%
  \setlength\@tempdima{\algorithmicindent}%
  \OldStatex\hskip\dimexpr#1\@tempdima\relax}

\definecolor{mygreen}{rgb}{0.64, 0.76, 0.68}
\definecolor{myyellow}{rgb}{0.98, 0.94, 0.75}
\definecolor{mygreen}{rgb}{0.68, 0.85, 0.9}
\definecolor{mypink}{rgb}{0.99, 0.87, 0.9}
\definecolor{myblue}{rgb}{0.82, 0.94, 0.75}
\def\bg{\mathbf{g}}

\title{Rethinking Generalization of Neural Models: \\A Named Entity Recognition  Case Study}
 
\newcommand*\samethanks[1][\value{footnote}]{\footnotemark[#1]}

\author{Jinlan Fu$\dag $\thanks{These two authors contributed equally}, Pengfei Liu$ \sharp$\samethanks, Qi Zhang$\dag$, Xuanjing Huang$\dag $\\
$\dag$School of Computer Science, Fudan University, \\
$\sharp$ Language Technologies Institute, Carnegie Mellon University, \\
\{fujl16,qz,xjhuang\}@fudan.edu.cn, pliu3@andrew.cmu.edu
 \\
}
 
\begin{document}

\maketitle

\begin{abstract}
While neural network-based models have achieved impressive performance on a large body of NLP tasks, the generalization behavior of different models remains poorly understood: Does this excellent performance imply a perfect generalization model, or are there still some limitations? In this paper, we take the NER task as a testbed to analyze the generalization behavior of existing models from different perspectives and characterize the differences of their generalization abilities through the lens of our proposed measures, which guides us to better design models and training methods. Experiments with in-depth analyses diagnose the bottleneck of existing neural NER models in terms of breakdown performance analysis, annotation errors, dataset bias, and category relationships, which suggest directions for improvement. \textbf{We have released the datasets: (\textit{ReCoNLL}, \textit{PLONER}) for the future research at our project page:} \url{http://pfliu.com/InterpretNER/}. \footnote{As a by-product of this paper, we have open-sourced a project that involves a comprehensive summary of recent NER papers and classifies them into different research topics: \url{https://github.com/pfliu-nlp/Named-Entity-Recognition-NER-Papers}.}.
\end{abstract}

\section{Introduction\label{sec:intro}}
 
Neural network-based models have achieved great success on a wide range of NLP tasks \cite{devlin2018bert,bahdanau2014neural}. However, the generalization behaviors of neural networks remain largely unexplained.
Recently, some researchers are beginning to realize this problem and attempt to understand the generalization behavior of neural networks in terms of network architectures or optimization procedure \cite{zhang2016understanding,baluja2017adversarial,schmidt2018adversarially}.
However, it is incomplete to ignore the characteristics of tasks and datasets for generalization analysis since it not only depends on the model's architectures but on the data itself \cite{arpit2017closer}.

In NLP, there is a massive gap between the growing task performance and the understanding of model generalization behavior. Many tasks have reached a plateau in the performance on a particular dataset \cite{rajpurkar2018know,devlin2018bert}, which calls for a data-dependent understanding of models' generalization behavior.

In this paper, we  take a step further towards
diagnosing and characterizing generalization in the context of a specific task.
Concretely, we take named entity recognition (NER) task as a study case and investigate three crucial yet rarely raised questions through \textit{entity}- and \textit{class}-centric  generalization analyses.

\textbf{Q1}: Does our model really have generalization ability, or it just pretends to understand and make some shallow template matches as observed in \cite{jia2017adversarial}?
We devise a measure, which can break down the test set into different interpretable groups, helping us diagnosing inadequacies in the generalization of NER models (Sec.~\ref{sec:Q1}).
Furthermore, this measure makes it easier to find human annotation errors, which cover the actual generalization ability of the existing models (Sec.~\ref{sec:Q1}).
\textbf{Q2}: What factor of a dataset can distinguish neural networks that generalize well from those that don\textquotesingle t? 
We introduce two metrics to quantify the dataset bias in a cross-dataset experimental setting, enabling us to better understand how the dataset bias influences the models' generalization ability (Sec.~\ref{sec:Q2}).
\textbf{Q3}: How does the relationship between entity categories influence the difficulty of model learning?
Our class-centric analysis shows that if two categories, e.g. $\mathscr{C}_1$ and $\mathscr{C}_2$, have overlaps (i.e. sharing a subset of entities), then most of the errors on $\mathscr{C}_1$ made by the model are due to mistakenly predicting $\mathscr{C}_1$ as $\mathscr{C}_2$ (Sec.~\ref{sec:Q3}). 
Our experiment results show the prospects for further gains for these problems from novel architecture design and knowledge pre-training seem quite limited (Sec.~\ref{sec:Q3}).
Tab.~\ref{tab:exp_design} shows the framework of our experimental designs.

\textbf{Main Contributions}
This paper understands the generalization behavior from multiple novel angles, which contributes from the following two perspectives:
1) For the task itself, we identify the bottleneck of existing methods on the NER task in terms of breakdown performance analysis, annotation errors, dataset bias, and category relationships, which suggest directions for improvement and can drive the progress of this area.
2) Other tasks can benefit from the research evidence found in this study.
For example, this paper not only shows that utilizing less but more relevant data can achieve better performance, but also provides an effective and principled way to select more relevant training samples.

\begin{table*}[htb]
  \centering \footnotesize
    \begin{tabular}{lcp{4cm}p{8cm}}
    \toprule
    \centering{\textbf{Views}} & \textbf{Q.}  & \centering{\textbf{Measures}} & \multicolumn{1}{c}{\textbf{Applications}  } \\
    \midrule
    \multirow{4}[4]{*}[-0cm]{Entity} & \multirow{2}[2]{*}[-0cm]{Q1 (Sec.~\ref{sec:Q1})}  & \multicolumn{1}{l}{\multirow{2}[2]{*}{Entity Coverage Ratio }} & (Exp I)   Breaking down the test set \\
          &       &       & (Exp-II) Annotation errors detecting and fixing\\   
\cmidrule{2-4}          & \multirow{2}[2]{*}{Q2 (Sec.~\ref{sec:Q2})} & \multicolumn{1}{l}{Expectation of Coverage Ratio} & (Exp-III)  Cross-dataset generalization  \\
          &       & \multicolumn{1}{l}{\multirow{1}{*}[-0cm]{Contextual Coverage Ratio}}  & (Exp-IV) Order matters for data augmentation  \\
    \midrule
    \multirow{2}{*}[-0cm]{Category} & \multirow{2}[2]{*}[-0cm]{Q3 (Sec.~\ref{sec:Q3})} & \multicolumn{1}{l}{\multirow{2}{*}[-0cm]{Consistency}} & (Exp V) Probing inter-category relationships \\
          &       &       & (Exp-VI) Exploring the errors of hard cases  \\
    \bottomrule
    \end{tabular}%
  \caption{Outline of our experiment designs. Q1: Does our model really have generalization? Q2: What factor of a dataset can distinguish neural networks that generalize well from those that don't? Q3: How does the relationship between entity categories influence the difficulty of model learning? } 
    \label{tab:exp_design}
\end{table*}

\textbf{Observations} Our findings are summarized as follows:
    (1) The fine-grained evaluation based on our proposed measure reveals that the performance of existing models (including the state-of-the-art model) heavily influenced by the degree to which test entities have been seen in training set \textit{with the same label} (Exp-I in Sec.~\ref{sec:Q1}).
    (2) The proposed measure enables us to detect human annotation errors, which cover the actual generalization ability of the existing model. We observe that once these errors are fixed, previous models can achieve new state-of-the-art results, $93.78$ F1-score on CoNLL2003, which refers to Exp-II in Sec.~\ref{sec:Q1}.     
    (3) We introduce two measures to characterize the data bias and the cross-dataset generalization experiment shows that the performance of NER systems is influenced not only by whether the test entity has been seen in the training set but also by whether the context of the test entity has been observed (Exp-III in Sec.~\ref{sec:Q2}).
    (4) Providing more training samples is not a guarantee of better results.
    A targeted increase in training samples will make it more profitable, which refers to Exp-IV in Sec.~\ref{sec:Q2}. 
   (5) The relationship between entity categories influences the difficulty of model learning, which leads to some hard test samples that are difficult to solve using common learning methods, which refer to Exp-V and Exp-VI in Sec.~\ref{sec:Q3}.

\section{Related Work}
Our work can be uniquely positioned in the context of the following two aspects.

\textbf{Neural Network-based Models for NER}
Some researchers design different architectures which vary in word encoder \cite{chiu2016named,ma2016end}, sentence encoder \cite{huang2015bidirectional,ma2016end,chiu2016named} and decoder (CRF) \cite{huang2015bidirectional}.
Some works explore how to transfer learned parameters from the source domain to a new domain \cite{chen2019transfer,lin2018neural,cao2018adversarial}. 
Recently, \cite{yang2018design,reimers2017optimal} systematically analyze neural NER models to provide useful guidelines for NLP practitioners.
Different from the above works, instead of exploring the possibility for a new state-of-the-art in this paper, we aim to bridge the gap between the growing task performance and the understanding of model generalization behavior.

\textbf{Analyzing Generalization Ability of Neural Networks}
Most existing works have analyzed the generalization power of DNNs by quantifying how the number of parameters, noise label, regularization, influence training process on a set of simple classification tasks.
\cite{fort2019stiffness} investigate neural network training and generalization by introducing a measure and study how it varies with training iteration and learning rate.
\cite{zhang2016understanding,arpit2017closer} explore the generalization of the neural network by showing how the impact of representational capacity changes with varying noise levels and regularization. 
The goal of this paper is to study in the light of a specific NLP task, discussing how neural networks achieve linguistic generalization abilities. 
More recently, \cite{zhong-etal-2019-closer}
try to analyze the generalization of neural networks in the text summarization task, which is mainly performed from a dataset perspective.

\section{Task, Methods, and Datasets}
\subsection{Task Description}

Named entity recognition (NER) is usually formulated as a sequence labeling problem \cite{borthwick1998exploiting}.
Formally, let $X = \{x_1,x_2,\ldots, x_T\}$ be an input sequence and $Y = \{y_1,y_2,\ldots,y_T\}$ be the output tags.
The goal of this task is to estimate the conditional probability:
$P(Y | X) = P(y_t|X,y_1,\cdots,y_{t-1})$

\paragraph{Why do We Choose the NER Task?}
The goal of this paper is to study how neural networks achieve linguistic-level generalization abilities via the lens of a well-chosen NLP task. 
Compared with other general classification tasks, the NER task is particularly suitable here because 1) it contains more category labels; 2) different categories contain a  number of training samples, which provides an ideal testbed for us to observe the generalization behavior of neural networks.
Although our focus is on NER tasks, our solution can be ported to the other tagging problems.

\subsection{Neural Network-based Methods for NER}

To evaluate the importance of different components of the NER systems, we varied our models mainly in terms of three aspects: different choices of character-, word-, and sentence-level encoders and decoders. 
Tab.\ref{tab:allmodels} illustrates the models we have studied in this paper. 
Specifically, for Exp-I, we mainly focus on how different choices of pre-trained models \cite{mikolov2013efficient,peters2018deep,devlin2018bert} influence systems' generalization abilities. All models adopt LSTM as sentence encoder and CRF as the decoder. 
For Exp-III and Exp-IV, we use \textit{CnoneWrandlstmCrf} model to achieve cross-dataset generalization evaluation.

For Exp-V, we adopt \textit{CcnnWglovelstmMLP} architecture since \textit{MLP} decoder is easy to compute the measure \textit{Consistency}.
For Exp-VI, detailed choices of evaluated models are listed in Tab.\ref{tab:error}.

\renewcommand\tabcolsep{2pt}
\begin{table}[!htb]
  \centering \scriptsize
 
    \begin{tabular}{l|ccccc|ccc|cc|cc}
    \toprule
    
   \multicolumn{1}{c|}{\multirow{2}[4]{*}{\textbf{Models}}}   & \multicolumn{5}{c}{\textbf{Character }} & \multicolumn{3}{|c}{\textbf{Word}} & \multicolumn{2}{|c}{\textbf{Sentence}} & \multicolumn{2}{|c}{\textbf{Decoder}} \\
\cmidrule{2-13}        & \textbf{\rotatebox{90}{none}} & \textbf{\rotatebox{90}{cnn}} & \textbf{\rotatebox{90}{elmo}} &  \textbf{\rotatebox{90}{flair}} & \textbf{\rotatebox{90}{bert}} &  \textbf{\rotatebox{90}{none}} & \textbf{\rotatebox{90}{rand}} & \textbf{\rotatebox{90}{glove}} & \textbf{\rotatebox{90}{lstm}} & \textbf{\rotatebox{90}{cnn}} & \textbf{\rotatebox{90}{crf}} & \textbf{\rotatebox{90}{mlp}} \\
    \midrule
    \textit{CcnnWglovecnnMlp} &     &  $\surd$     & & &       &      &       &  $\surd$      &     & $\surd$    & & $\surd$           \\
    \textit{CcnnWglovelstmMlp} &    &  $\surd$   & &   &       &      &       &  $\surd$      & $\surd$    &        &    &  $\surd$        \\
    \textit{CcnnWglovecnnCrf} &      &  $\surd$  & &    &       &      &       &  $\surd$      &     & $\surd$       & $\surd$    &   \\
    \textit{CnoneWglovelstmCrf} &  $\surd$  & &  &     &       &      &       &  $\surd$      & $\surd$     &       & $\surd$    &   \\
    \textit{CcnnWglovelstmCrf} &       & $\surd$   & &   &       &       &       &   $\surd$       & $\surd$       &      & $\surd$      &       \\
    \textit{CelmWnonelstmCrf} &       &      &  $\surd$  & &    & $\surd$   &       & & $\surd$      &       & $\surd$      &       \\
    \textit{CelmWglovelstmCrf} &       &       & $\surd$   & &   &       &       & $\surd$      &  $\surd$      &      & $\surd$      &       \\
     \textit{CnoneWrandlstmCrf} &  $\surd$    & &  &      &       &      &     $\surd$ &         & $\surd$     &       & $\surd$    &   \\
\textit{CflairWglovelstmCrf} &     & &  &$\surd$       &       &      &    &  $\surd$         & $\surd$    &    & $\surd$  &   \\
\textit{CbertWnonelstmCrf} &     & &  &       & $\surd$      &      &    &  $\surd$         & $\surd$    &    & $\surd$  &   \\

    \bottomrule
    \end{tabular}
     \caption{Neural NER systems with different architectures and pre-trained knowledge, which we studied in this paper.} 
  \label{tab:allmodels}
\end{table}

\subsection{NER Datasets for Evaluation}
We conduct experiments on three benchmark datasets: the CoNLL2003 NER dataset, the WNUT16 dataset, an d  OntoNotes 5.0 dataset. 
The CoNLL2003 NER dataset \cite{sang2003introduction} is based on Reuters data \cite{collobert2011natural}. WNUT16 dataset is provided by the second shared task at WNUT-2016. The OntoNotes 5.0 dataset \cite{weischedel2013ontonotes} is collected from telephone conversations (TC),
newswire (NW), newsgroups, broadcast news (BN), broadcast conversation (BC) and weblogs (WB), pivot text (PT) and magazine genre (MZ). Due to the lack of NER labels of PT and the insufficient amount of data of TC, we only evaluate the other five domains. 

\section{Experiment and Analysis}

\subsection{Diagnosing Generalization with \textit{Entity Coverage Ratio}} \label{sec:Q1}

The generalization ability of neural models is often evaluated based on a holistic metric over the whole test set.
For example, the performance of the NER system is commonly measured by the F1 score.
Despite its effectiveness, this holistic metric fails to provide fine-grained analysis and as a result, we are not clear about what the strengths and weaknesses of a specific NER system are.

Driven by \textbf{Q1}, we propose to shift the focus of evaluation from a holistic way to fine-grained way, navigating directly to the parts which influence the generalization ability of neural NER models.
We approach the above end by introducing the notion of \textit{entity coverage ratio} (ECR) for each test entity, by which the test set will be divided into different sub-sets, and the overall performance could be broken down into interpretable categories.

\begin{table}[!htbp]
\small
\begin{tabular}{lp{6.3cm}}
\toprule
$\rho(e)$ & \textbf{Interpretation}  \\
\midrule
$\rho=1$    & Entity $e$ appears in train set with only label $k$           \\
$\rho \in (0,1)$    & Entity $e$ appears in train set with diverse labels         \\
$\rho=0 \wedge C\neq0$     & Entity $e$ appears in train set but without label $k$  \\
$\rho=0 \wedge C=0$     & Entity $e$ doesn't appear in train set \\
\bottomrule
\end{tabular}
\caption{Interpretation of $\rho$ with different values.} \label{tab:exp-rho}
\end{table}

\renewcommand\tabcolsep{8pt}
\begin{table*}[!htb] 
  \centering \footnotesize
    \begin{tabular}{lll|cccccc}
    \toprule
    \multirow{2}[4]{*}{\textbf{Datasets}} & \multicolumn{2}{c|}{\textbf{Embed-layer}} & 
    \multicolumn{6}{c}{\textbf{Entity Coverage Rate}} \\
\cmidrule{2-9}          & \textbf{Char} & \textbf{Word} & 
$\mathbf{Overall}$ & $1$ & ($0.5,1$) & ($0,0.5$] & $C\neq0$ & $C=0$ \\
    \midrule
    \multirow{8}[2]{*}{CoNLL} 
    & CNN  & -      & 76.42   & 79.94 & 86.99 & 78.84   & 69.74 & 77.61\\
    & FLAIR  & -     & 89.98  & 95.30 & 95.58 & 82.39 & 72.16 & 90.39\\
         & ELMo  & -     & 91.79  & 97.61 & 95.98 & 85.15  & 71.43 & 92.22\\
          & BERT  & -    & 91.34 & 97.72 & 95.17 & 86.66 & \textbf{77.83} & 92.37 \\
          
          & -  & Rand      & 78.43 & 95.05 & 94.75 & 73.54 & 37.97 & 66.40 \\
          & -  & GloVe      & 89.10  & 98.44 & 96.31 & 81.34  & 57.80 & 87.23\\
          & CNN  & Rand      & 82.88  & 94.13 & 94.48 & 74.25  & 47.78 & 78.91\\
          & CNN & GloVe      & 90.33   & 98.32 & 95.94 & 80.33  & 59.67 & 89.74\\
          & ELMo & GloVe      & 92.46 & 98.08  & \textbf{96.46} & 86.14 & 69.79 & 93.08 \\
          & FLAIR & GloVe      & \textbf{93.03} & \textbf{98.56}  & 96.38 & \textbf{87.07} & 73.58 & \textbf{93.42} \\

\cmidrule{1-9}    \multirow{10}[2]{*}{WNUT} 
& CNN  & -      & 20.88  & 45.99 & 67.01 & 40.25  & 19.14 & 19.74 \\
& FLAIR  & -     & 41.49  & 81.15 & 88.14 & 54.36 & 39.56 & 43.44\\
& ELMo  & -     & 43.70  & 88.72 & 90.83 & 55.56 & \textbf{44.19} & 43.32\\

& BERT  & -     & 44.08 & 77.75 & 81.61 & 49.74 & 34.65 & 41.92 \\
& -  & Rand     & 14.97 & 60.62 & 83.84 & 50.00 & 3.90 & 4.77 \\
& -  & GloVe      & 37.28  & 89.29 & 92.62 & 45.65  & 35.34 & 35.15\\
& CNN  & Rand     & 22.29 & 48.88 & 71.43 & 39.08 & 16.75 & 18.83 \\
& CNN  & GloVe      & 40.72 & 86.12 & \textbf{92.24} & 49.74 & 26.67 & 40.06 \\
& ELMo  & GloVe & 45.33 & 90.38 & 89.92 & 56.57 & 37.8 &46.58\\
& FLAIR  & GloVe & \textbf{45.96} & \textbf{90.52} & 89.92 & \textbf{61.69} & 42.07 &\textbf{48.38}\\
    \bottomrule
    \end{tabular}

 \caption{
 The breakdown performance on CoNLL and WNUT datasets with different pre-training strategies, which is based on the LSTM as sentence encoder and CRF as the decoder. 
 ``\texttt{Rand}'' represents the word representations are randomly initialized.
 ``\texttt{Overall}'' denotes the F1 score on the whole test set and the names of the last five columns correspond to $\rho$ definition in Tab.\ref{tab:exp-rho}. 
   \label{tab:allreslts}
 }
\end{table*}

\paragraph{Entity Coverage Ratio (ECR)}
The measure entity coverage ratio is used to describe the degree to which entities in the test set have been seen in the training set with the same category. 
Specifically, we refer to $e_i$ as a test entity, whose coverage ratio is defined as:
\begin{equation}
    \rho(e_i) =
  \begin{cases}
  0 & \mbox{$C=0$ }\\
  (\sum_{k=1}^K \frac{\#(e_i^{tr,k})}{C^{tr}} \dot \#(e_i^{te,k})) / C^{te}
   &\mbox{otherwise}
  \end{cases} \label{eq:def-rho}
\end{equation} 
where $e_i^{tr,k}$ is the entity $e_i$ in the training set with ground truth label $k$, $e_i^{te,k}$ is the entity $e_i$ in the test set with ground truth label $k$, $C^{tr} = \sum_{k=1}^K \#(e_{i}^{tr,k})$, $C^{te} = \sum_{k=1}^K \#(e_i^{te,k})$, and $\#$ denotes the counting operation. 

For example, in the training set, ``\texttt{chelsea}'' is labeled as the category \texttt{Person} 6 times, and \texttt{Organization} 4 times, while in the test set, labeled as \texttt{Person} 3 times and \texttt{Organization} 2 times, so $\rho$ (``\texttt{chelsea}'') $=(0.6\times3+0.4\times2) / 3 = 0.52$.
According to Eq.\ref{eq:def-rho},
we can investigate the relationship between the coverage ratio of the entity $e_i$ and model's generalization ability on this entity.
The possible values of $\rho(e_i)$ and their corresponding interpretation can be found in Tab.~\ref{tab:exp-rho}.

\paragraph{Exp-I: Breaking Down the Test Set}  \mbox{} \\

\label{p:exp-I}
Instead of utilizing a holistic metric on the whole dataset, we break down the test set into interpretable regions by the measure $\rho$ and then observe how the generalization ability of the NER models varies with it.

\paragraph{Results}

Based on Tab.~\ref{tab:allreslts} and driven by \textbf{Q1}, our observations are:
1) In general, the part of test entities with high performance are usually the ones that appear in the training set. By contrast, if the test entity is unseen, it will achieve a lower performance.
2) No matter what level (character or word) pre-trained embeddings are introduced, the performances of unseen entities are largely improved.
3) Comparing two different levels of pre-trained methods, 
\texttt{ELMo} and \texttt{FLAIR} achieve better performances on unseen entities but have not shown significant gain on seen entities.
4) Compared to \texttt{Rand}, \texttt{CNN} shows its superior performance on the prediction of unseen entities.
5) For different parts of the test set, we find $C\neq0$ is the most challenging part (even for the state-of-the-art model), followed by $C=0$ and ($0,0.5$].
Interestingly, on the CoNLL dataset, we find that if the test entity is labeled as a different category in the training set, it will be more difficult to learn compared with entities which have not been seen in the training set.
6) \textit{We find that the character- and the word-level pre-trained embeddings are complementary to each other}. Combining these two types of \textit{pre-trained} knowledge will further improve the performance by a considerable margin. 

\paragraph{Exp-II: Annotation Errors Detecting and Fixing}  \mbox{} \\ 

For each test entity with tag $k$, the measure \textit{ECR} quantifies its \textit{label ambiguity}: the proportion that this entity is labeled as $k$ in the training set.
Its intriguing property could help us find the annotation errors of the dataset.

\paragraph{Detecting Errors}
Specifically, since $\rho$ measures the degree to which entities in the test set have been seen in the training set with the same label, 
the value of $\rho$ within some ranges suggests that corresponding entities are more prone to annotation errors, such as $\rho=0, C \neq 0$ (entity $e^k$ appeared in train set but without label $k$) and $\rho \in (0,0.5]$ (entity $e^k$ appeared in train set with diverse labels). 

\paragraph{Fixing Errors}

While researchers have been aware of annotation errors, such as on the tasks of Part-of-Speech \cite{manning2011part} and Chinese word segmentation \cite{ma2018state}, yet few attempts have been made to fix them.
The significance of correcting annotation errors for tagging tasks has been originally mentioned by \cite{manning2011part}.
In this paper, we argue that fixing annotation errors can not only boost the NER performance,
but can reflect the true generalization ability of the existing models, making it possible to identify the real weaknesses of current systems.

\paragraph{Evaluation on Revised CoNLL (\textit{ReCoNLL})}

Many errors and inconsistencies in NER datasets are quite non-systematic and are hard to fix by deterministic rules. Therefore, we manually fixed errors with the instruction of the measure ECR (entity coverage ratio).
Finally, we corrected 65 sentences in the test set, and 14 sentences in training set.
When the revised dataset is ready, we re-train several typical NER models and make a comparison to the original ones. 

The results are shown in Tab.~\ref{tab:revised_conll}.
We find that once these errors are fixed, the performance of all these models has been improved, which indicates that human annotation errors cover the actual generalization ability of the existing model. Notably, the NER model \textit{FLAIR} has driven the state-of-the-art result to a new level.

\renewcommand\tabcolsep{4pt}
\begin{table}[htbp]
  \centering \footnotesize 
    \begin{tabular}{lrc}
    \toprule
    \textbf{Model} & \multicolumn{1}{l}{\textbf{CoNLL}} & \multicolumn{1}{l}{\textbf{ReCoNLL}} \\
    \midrule
    \cite{devlin2018bert} &  92.80     & - \\
    \cite{peters2018deep} &    92.22   & - \\
    \cite{akbik2018contextual} &  93.09   & - \\
    \cite{akbikpooled} &  \textbf{93.18}   & - \\
    \midrule
    Our Implementation & & \\
    \midrule    
    GloVe &     89.10  & 89.85 \\
    ELMo  &     91.79  &  92.65\\
    BERT  &     91.34  & 92.16 \\
    FLAIR  & 93.03 & \textbf{93.78} \\
    \bottomrule
    \end{tabular}%
      \caption{The test performance (F1 score) on CoNLL 2003 and its revised version. 
      }
  \label{tab:revised_conll}
\end{table}

\subsection{Measuring Dataset Bias} \label{sec:Q2}
To answer the question \textbf{Q2}: ``what factor of a dataset can distinguish neural networks that generalize well from those that don't'',
in this section, we introduce two measures, which can quantify the relationship of entities between training and test sets from dataset-level and help us understand the generalization behavior.

\paragraph{Expectation of Entity Coverage Ratio (EECR)}
Here, we define the expectation of the coverage ratio over all entities in test data as $E_{\rho}(e)$ as follows:
\begin{align}
    E_{\rho}(e) = \sum_{i \in N_e}\rho(e_i) * \mathrm{freq}(e_i)
\end{align}
in which $N_e$ denotes the number of unique test entities and $\mathrm{freq}(e_i)$ represents the frequency of the test entity $e_i$.

This index measures the degree to which the test entities have been seen in the training set. A higher value is suggestive of a larger proportion of entities with high coverage ratio.

\paragraph{Contextual Coverage Ratio (CCR)}
We introduce a notion of $\eta$ to measure the contextual similarity of entities belonging to the same category but from the training and the test sets, respectively. 

\begin{align} \small
    \eta^{k}(D_{tr},D_{te}) = 
    \sum_{f_i \in \phi^k_{te} } \sum_{f_j \in \phi^k_{tr} } p_{f_i} p_{f_j} \mathrm{Sim}(v_{f_i},v_{f_j})
\end{align}
where $k$ denotes the category of an entity. $D_{tr}$ and $D_{te}$ represents the training and test sets.
$\phi_{tr}^{k}$ denotes a set of the high-frequency contextual patterns in which entities in training set reside in. We set the window size to 3, and  choose 30 bigrams and 20 trigrams, then we obtain their vector representation $v_{f_{*}}$ of each word span using BERT followed by a mean operation. $\mathrm{Sim}(\cdot)$ is a cosine-similarity function.
$p_{f_i}$ is the probability of the contextual pattern $f_i$, which is using the frequency of the contextual pattern divided by the total contextual patterns' frequency.

\paragraph{Exp-III: Cross-dataset Generalization} \mbox{} \\ \label{p:aedf}

The Expectation of Entity Coverage Ratio and Contextual Coverage Ratio can measure the similarity between training and test set from a different perspective.
Next, we show how these two measures correlate with the model's performance by a cross-dataset generalization experiment.
\paragraph{Data Construction: \textit{PLONER}}
We re-purpose a dataset for  cross-domain generalization evaluation, in which three types of entities (\textsc{Person}, \textsc{Location}, \textsc{Organization}) from different domains are involved, therefore named ``\textbf{PLO}NER'' dataset.
Specifically, we pick a set of representative NER datasets including: \texttt{WNUT16}, \texttt{CoNLL03}, \texttt{OntoNotes-bn}, \texttt{OntoNotes-wb}, \texttt{OntoNotes-mz}, \texttt{OntoNotes-nw}, and \texttt{OntoNotes-bc}.
These datasets use disparate entity classification schemes, which makes it hard to conduct zero-shot transfer. We collapse types into standard categories used in the MUC \cite{grishman1996message} competitions (\textsc{Person}, \textsc{Location}, \textsc{Organization}) and the other categories are dropped.
\footnote{We have released the dataset.}
To be fair, we limited the number of samples in each dataset to the same 2,500.

\renewcommand\tabcolsep{4pt}
\begin{table}[htb]
  \centering  \scriptsize 
    \begin{tabular}{cl|ccccccc|c}
    \toprule
   \textbf{Matrix} & \textbf{Train} & \textbf{WN.} & \textbf{Co.} & \textbf{BN} & \textbf{WB} & \textbf{MZ} & \textbf{NW} & \textbf{BC}  & \textbf{P-row}\\
    \midrule
    \multirow{7}[2]{*}{\rotatebox{90}{$M_{F1}$}} & WN.  & \textbf{46.6} & 16.7 & 12.0 & \fcolorbox{blue}{white}{13.8} & 11.4  & 6.50   & 11.7  &  \multicolumn{1}{>{\columncolor{myblue}}l}{0.98} \\
    & Co. & 22.2 & \textbf{70.2} & 19.4 & \fcolorbox{blue}{white}{17.5} & 12.7 & 19.8 & 17.3 &\multicolumn{1}{>{\columncolor{myblue}}l}{0.93}\\
    & BN & 24.4 & 35.2 & \textbf{65.7} & 35.4 & 28.6 & 36.2  & 42.3 &\multicolumn{1}{>{\columncolor{myblue}}l}{0.93}\\
    & WB & 19.6 & 25.6 & 28.0 & \textbf{55.7} & 17.3 & 28.3 & 25.8 &\multicolumn{1}{>{\columncolor{myblue}}l}{0.83}\\
    & MZ & 15.3 & 25.0 & 29.7 & \fcolorbox{red}{white}{21.8}  & \textbf{67.8} & 32.2 & 26.6 &\multicolumn{1}{>{\columncolor{myblue}}l}{0.89}\\
    & NW & 22.2 & 24.2  & 32.6 & 29.3 & 24.6 & \textbf{68.4} & 26.9 &\multicolumn{1}{>{\columncolor{myblue}}l}{0.87}\\
    & BC & 28.6 & 29.1 & 41.1 & 43.6 & 25.5 & 33.2 & \textbf{70.3} &\multicolumn{1}{>{\columncolor{myblue}}l}{0.90}\\
    \midrule
   &\multicolumn{1}{>{\columncolor{myblue}}l}{P-col}  &\multicolumn{1}{>{\columncolor{myblue}}l}{0.93} &\multicolumn{1}{>{\columncolor{myblue}}l}{0.97} &\multicolumn{1}{>{\columncolor{myblue}}l}{0.97} &\multicolumn{1}{>{\columncolor{myblue}}l}{0.83} &\multicolumn{1}{>{\columncolor{myblue}}l}{0.96} &\multicolumn{1}{>{\columncolor{myblue}}l}{0.94}  &\multicolumn{1}{>{\columncolor{myblue}}l}{0.96} &\multicolumn{1}{>{\columncolor{mygreen}}l}{0.88}\\
    \midrule
    \multirow{7}[2]{*}{\rotatebox{90}{$M_{\rho}$}} & WN.  & \textbf{1.00}     & 0.26 & 0.29 & \fcolorbox{blue}{white}{0.21} & 0.07 & 0.24 & 0.31 & \multicolumn{1}{>{\columncolor{mypink}}l}{0.96}\\
          & Co. & 0.325 & \textbf{1.00}     & 0.35 & \fcolorbox{blue}{white}{0.37} & 0.29 & 0.36 & 0.37 &\multicolumn{1}{>{\columncolor{mypink}}l}{0.94}\\
          & BN & 0.35 & 0.33 & \textbf{1.00}     & 0.49 & 0.32  & 0.45 & 0.65 &\multicolumn{1}{>{\columncolor{mypink}}l}{0.95} \\
          & WB & 0.27 & 0.22  & 0.43 & \textbf{1.00}     & 0.29 & 0.34 & 0.52 &\multicolumn{1}{>{\columncolor{mypink}}l}{0.91}\\
          & MZ & 0.15 & 0.15 & 0.24 & \fcolorbox{red}{white}{0.15} & \textbf{1.00}     & 0.31 & 0.23 &\multicolumn{1}{>{\columncolor{mypink}}l}{0.99}\\
          & NW & 0.20 & 0.27  & 0.38 & 0.34 & 0.34 & \textbf{1.00}     & 0.41 &\multicolumn{1}{>{\columncolor{mypink}}l}{0.92}\\
          & BC & 0.28 & 0.22 & 0.52 & 0.49 & 0.35 & 0.36  & \textbf{1.00} &\multicolumn{1}{>{\columncolor{mypink}}l}{0.87}\\
    \midrule
   &\multicolumn{1}{>{\columncolor{mypink}}l}{P-col} &\multicolumn{1}{>{\columncolor{mypink}}l}{0.95} &\multicolumn{1}{>{\columncolor{mypink}}l}{0.95} &\multicolumn{1}{>{\columncolor{mypink}}l}{0.91} &\multicolumn{1}{>{\columncolor{mypink}}l}{0.89} &\multicolumn{1}{>{\columncolor{mypink}}l}{0.97} &\multicolumn{1}{>{\columncolor{mypink}}l}{0.92} &\multicolumn{1}{>{\columncolor{mypink}}l}{0.92} &\multicolumn{1}{>{\columncolor{mygreen}}l}{0.78}\\
    \midrule
    \multirow{7}[2]{*}{\rotatebox{90}{$M_{\phi}$}} & WN.  & \textbf{1.00}     & 0.06 & 0.08 & \fcolorbox{blue}{white}{0.11} & 0.07 & 0.03 & 0.12 &\multicolumn{1}{>{\columncolor{myyellow}}l}{0.99}\\
          & Co. & 0.176 & \textbf{1.00}     & 0.20 & \fcolorbox{blue}{white}{0.35} & 0.20   & 0.13 & 0.27 &\multicolumn{1}{>{\columncolor{myyellow}}l}{0.91}\\ 
          & BN & 0.19 & 0.22 & \textbf{1.00}     & 0.56 & 0.34 & 0.18 & 0.65 &\multicolumn{1}{>{\columncolor{myyellow}}l}{0.88}\\
          & WB & 0.29 & 0.26 & 0.41 & \textbf{0.88} & 0.42 & 0.23 & 0.70 &\multicolumn{1}{>{\columncolor{myyellow}}l}{0.70}\\
          & MZ & 0.32 & 0.28 & 0.43 & \fcolorbox{red}{white}{1.00}     & \textbf{1.00}     & 0.33 & 0.68 &\multicolumn{1}{>{\columncolor{myyellow}}l}{0.57}\\
          & NW & 0.36 & 0.22 & 0.42 & 0.90 & 0.53 & \textbf{1.00} & 0.69 &\multicolumn{1}{>{\columncolor{myyellow}}l}{0.70}\\
          & BC & 0.26 & 0.22 & 0.40 & 0.52 & 0.32 & 0.19 & \textbf{1.00}& \multicolumn{1}{>{\columncolor{myyellow}}l}{0.90}\\
    \midrule
  &\multicolumn{1}{>{\columncolor{myyellow}}l}{P-col}  &\multicolumn{1}{>{\columncolor{myyellow}}l}{0.85} &\multicolumn{1}{>{\columncolor{myyellow}}l}{0.97} &\multicolumn{1}{>{\columncolor{myyellow}}l}{0.96} &\multicolumn{1}{>{\columncolor{myyellow}}l}{0.45} &\multicolumn{1}{>{\columncolor{myyellow}}l}{0.93} &\multicolumn{1}{>{\columncolor{myyellow}}l}{0.93} &\multicolumn{1}{>{\columncolor{myyellow}}l}{0.83} &\multicolumn{1}{>{\columncolor{mygreen}}l}{0.89}\\
    
    \bottomrule
    \end{tabular}
    \caption{Illustration of F1 score, EECR, and CCR on cross-dataset setting. 
    \texttt{P-row} and \texttt{P-col} represent row- and column-wise Pearson correlation coefficient. \texttt{Green}, \texttt{Pink} and \texttt{Yellow} regions denote the correlation between $M_{F1}$ and  $M_{\rho}$+$M_{\phi}$, $M_{\rho}$, $M_{\phi}$ respectively. The \texttt{Blue} is the overall correlation coefficient. 
    }
  \label{tab:cross-data}
\end{table}

\paragraph{Results}
Tab.~\ref{tab:cross-data} shows the cross-dataset expectation of coverage ratio ($M_{\rho}$), contextual coverage ratio ($M_{\phi}$), and F1 score ($M_{F1}$). Each column corresponds to the performance when testing on one dataset and training on each of other datasets.
We detail our findings as follows:

1)    The diagonal elements of the $M_{F1}$ achieve the highest values, which suggests that models generalize poorly on the samples from different distributions (domains).

2)    The highest values are also achieved on the diagonal in $M_{\rho}$ and $M_{\phi}$. Additionally, from the values of Pearson coefficient, we could find the two measures: expectation of coverage ratio ($M_{\rho}$), contextual coverage ratio ($M_{\phi}$) correlate closely with F1-score $M_{\rho}$.

3) Column-wisely, given a test dataset,  $\rho$, $\phi$, and $F1$ can usually achieve the highest values on the same training set, which suggests we can select the most useful training sets through the measures $\rho$ and $\phi$ when
the distribution to be tested is given and we have some samples from it as the validation set.

4)    Given a test set, the training set with higher EECR (Expectation of Entity Coverage Ratio) value could also obtain a lower F1 score, since \textbf{entity coverage ratio is not the only factor that effects generalization and the contextual coverage ratio also matters}.

A significant case can be found in Tab.~\ref{tab:cross-data} (numbers in boxes), taking the \texttt{WB} as a test set, we observe that \texttt{WNUT} and \texttt{CoNLL} have higher ECR($\rho$) value than \texttt{MZ} while obtaining lower $F1$ score.
We can speculate the reason from the $\phi$-$M$, that the contextual coverage ratio between \texttt{WB} and \texttt{MZ} is much higher than utilizing \texttt{WNUT} and \texttt{CoNLL} as training sets.  
The above results show that \textit{the generalization ability of NER models is influenced not only by whether the test entity has been seen in the training set but also by whether the context of the test entity has been seen.}

\paragraph{Exp-IV: Order Matters for Data Augmentation} \mbox{} \\

The measure ECCR can be used to quantify the importance of different source domains, therefore allowing us to select suitable ones for data augmentation.
Next we will show how to utilize the ECCR metric to make better choices of source domains from the seven candidates: \texttt{WNUT16}, \texttt{CoNLL03}, \texttt{OntoNotes-bn}, \texttt{OntoNotes-wb}, \texttt{OntoNotes-mz}, \texttt{OntoNotes-nw}, and \texttt{OntoNotes-bc}.
We take \texttt{WNUT} as the tested object and continuously increase the training samples of above seven datasets in three ways:
1)  random order of EECR scores;
2)	descending order of EECR scores; 
3)	ascending order of EECR scores;

\begin{figure}[t]
    \centering
    \includegraphics[width=0.25\textwidth]{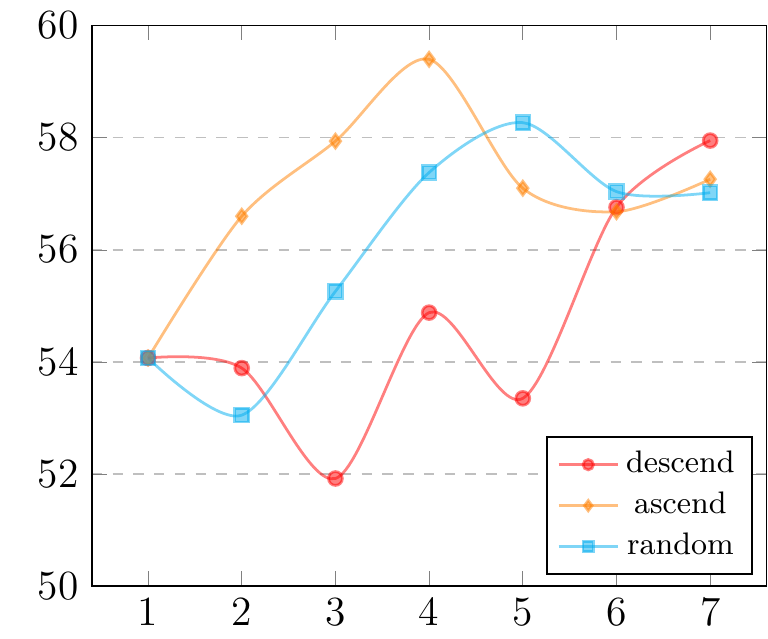}
    \caption{Changes of F1-score as more source domains are introduced in three different orders: descending order (red), ascending order (orange) and random order (blue) of EECR scores.}
    \label{fig:rho_exp}
\end{figure}

\paragraph{Results}
Fig.~\ref{fig:rho_exp} shows the results and we can find that
\textit{it is not that the more training data we have, the better performance we will obtain}.
When we introduce multiple training sets for data augmentation, the order of the distance between training sets and validation sets can help us select the most useful training sets.

\subsection{Diagnosing Generalization with \textit{Consistency}} \label{sec:Q3}
Above entity-centric analyses encourage us to find interpretable factors that affect the model's generalization ability.
We can also understand the models' generalization behavior from the perspective of the category of each entity.
To answer the question \textbf{Q3}, we propose to use a proxy measure \textit{consistency} via the angle of gradients to investigate how the relationship between entity categories influence the difficulty of model learning. 

The core idea behind the measure \textit{consistency} is to quantify the effect of different test samples on the trained parameters in NER models.
Formally, given a training sample $x$ and its ground truth label $y$,
we refer to $f(x,\theta)$ as the parameterized neural network.
Generally, the loss function $\mathcal{L}({f(x,\theta),y})$ shows the difference between model's output and ground truth label.
And the gradients of the loss with respect to $\theta$ can be formulated as: $\mathbf{g} = \nabla_{\theta}\mathcal{L}({f(x,\theta),y})$
Here, we propose to characterize the generalization ability of neural networks by observing the gradients' behaviors of test samples.
Specifically, given any two samples, the measure consistency $\mathrm{Cs}$ can be defined as the cosine angle of their gradients: $\mathrm{Cs}(\bg_1,\bg_2) = \frac{\bg_1\cdot \bg_2}{|\bg_1||\bg_2|}$, 
where $|\mathbf{v}|$ denotes the $L2$-norm of vector $\mathbf{v}$. $\bg_1$ and $\bg_2$ represent two gradient vectors derived from two samples.
The idea of utilizing the angle of gradients induced by two test examples has been originally explored on image classification \cite{fort2019stiffness}. Here, we extend this idea to NLP tasks.

\begin{algorithm}[t]
\caption{Consistency calculation and evaluation for Named Entity Recognition}
\label{alg:metanet}
\begin{algorithmic}[1]
{\footnotesize
\Require{Training dataset $\mathcal{D}^{tr}$ and multiple subsets of validation data $\mathcal{D}^{val}_1, \cdots, \mathcal{D}^{val}_K \subset \mathcal{D}^{val} $. }
\Comment{\textcolor[rgb]{0.00,0.59,0.00}{K is the number of categories}}
\Require{Parameters of the model $\theta \leftarrow \theta^{0}$}
\State{Train the model using $\mathcal{D}^{tr}$: $\theta \leftarrow \hat{\theta}$}
    \For{$p \in \lbrace 1 \cdots  K \rbrace$}
    \For{$q \in \lbrace 1 \cdots  K \rbrace$}
        \State{ Compute class-level consistency:
        $\delta(p,q) =  \frac{1}{|\mathcal{D}_p^{val}||\mathcal{D}_q^{val}|} \sum_{i}^{|\mathcal{D}_p^{val}|} \sum_{j}^{|\mathcal{D}_q^{val}|} \mathrm{Cs}(\bg_{e_i}^{p},\bg_{e_j}^{q})$
        \Comment{\textcolor[rgb]{0.00,0.59,0.00}{Eq.\ref{eq:delta}}}
        }
    \EndFor
    \EndFor
\Return{A consistency matrix $M \in \mathcal{R}^{K \times K}$}
}\end{algorithmic} 
  \label{alg:cs}
\end{algorithm}

\paragraph{Consistency Evaluation for NER}

Formally, given an entity $e^k$ and its label $y^k$, we refer to $\mathbf{g}_e^k = \nabla_{\theta}\mathcal{L}({f(x,\theta),y^k})$ as its generated gradient vector, where $x$ is the input sample containing entity $e^k$.
Then, for any two samples that contain two entities ($e_i$ and $e_j$) with different categories ($p$ and $q$),
we introduce the measure $\delta(p,q)$ to quantify the difference between two directions along which the parameters are updated.
\begin{align}
    \delta(p,q) = \frac{1}{C_pC_q} \sum_{i}^{C_p} \sum_{j}^{C_q} \mathrm{Cs}(\bg_{e_i}^{p},\bg_{e_j}^{q}) \label{eq:delta}
\end{align}
where $p$ and $q$ denote different entity categories. $C_p$ represents the number of test entities with category $p$.
Alg.~\ref{alg:cs} illustrates the process for consistency calculation and evaluation.

\paragraph{Exp-V: Probing Inter-category Relationships via \textit{Consistency}}
\mbox{} \\

\begin{figure}[!htbp]
    \centering
    \includegraphics[width=0.48\textwidth]{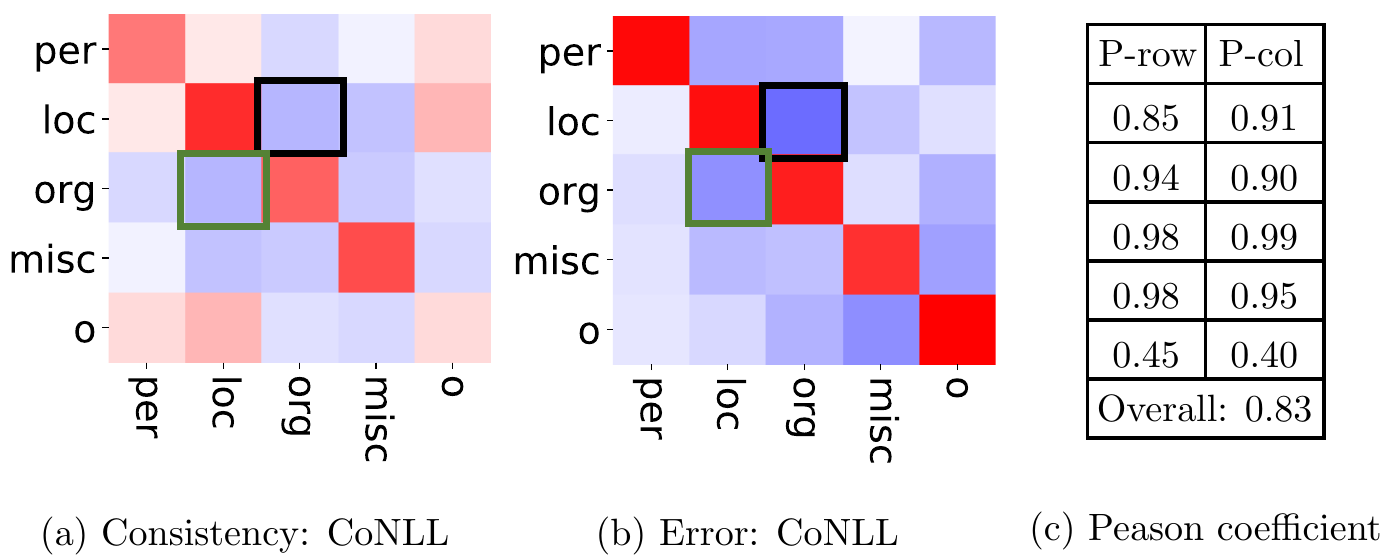}
    \caption{The alignment between \textit{consistency} $\mathrm{Cs}$ and the error ratio. Sub-figure (a) is the category-membership dependence of \textit{consistency} $\mathrm{Cs}$. Sub-figure (b) is the cross-category error ratios. 
    Sub-figure (c) denotes the Pearson coefficients between consistency values and error values.
    The change in color from blue to red represents the change in value from negative to positive.
    }
    \label{fig:cs-error}
\end{figure}

Given an NER model, we can understand its generalization ability by calculating the consistency matrix based on Alg.~\ref{alg:cs}.
As shown in Fig.~\ref{fig:cs-error}, the sub-figure (a) illustrates the \textit{consistency} matrice of two NER models trained on \texttt{CoNLL}  
As expected, the on-diagonal elements of $M_{p,q}$ $ (p=q)$ usually stay high, since it is easier for the model to find shared features between different entities within the same category.
Algorithmically speaking, a gradient step taken with respect to one test entity can reduce the loss on another test entity.

\begin{figure}[t]
  \centering
  \subfigure[The concept of \textit{consistency} $\mathrm{Cs}$]{
    \label{fig:consistency}
    \includegraphics[width=0.35\textwidth]{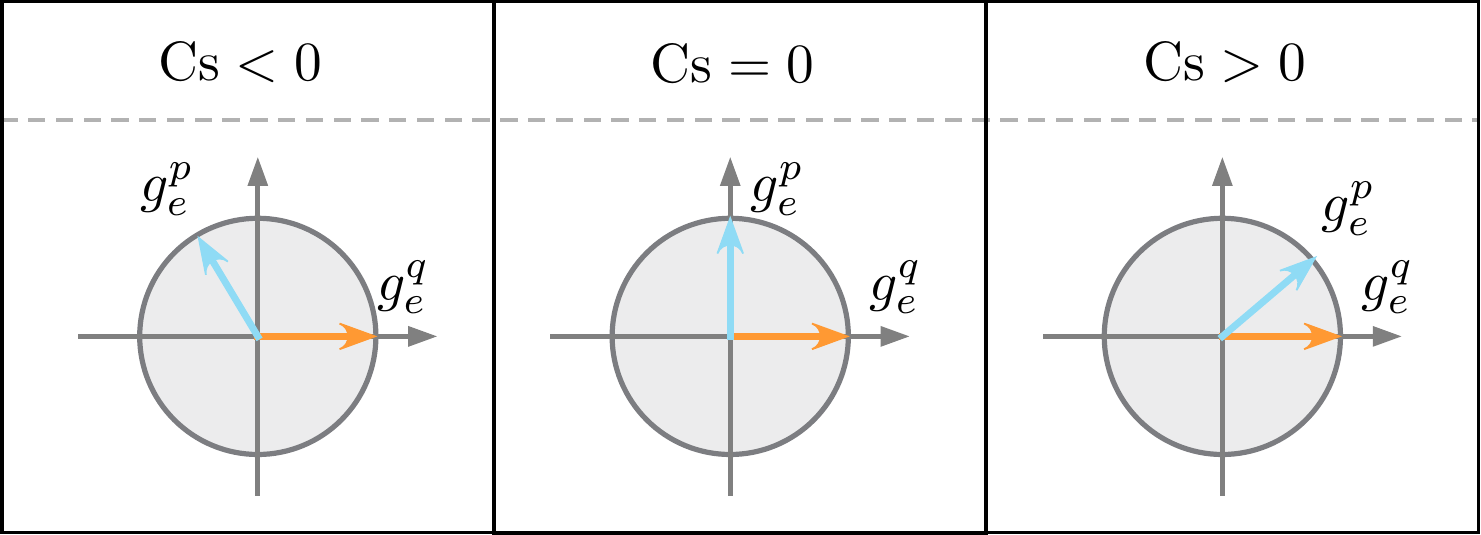}
  }
  \\
  \subfigure[Entity categories relationship ]{
    \label{fig:category}
    \includegraphics[width=0.35\textwidth]{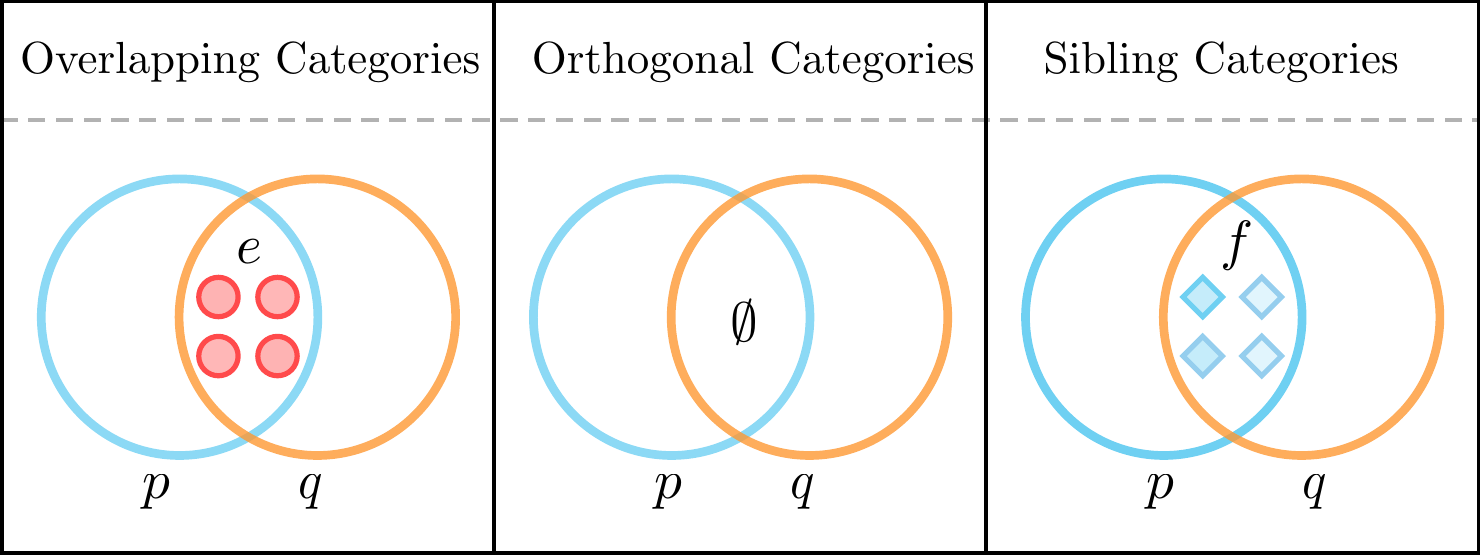}
  }
 \caption{Illustration of the concept of \textit{consistency} $\mathrm{Cs}$ and entity categories relationship. $g_{e}^{p}$ and $g_{e}^{q}$ represents the gradient of entity belongs to category $p$ and $q$, respectively. $e$ and $f$ are the collection of entities and features overlapped by categories $p$ and $q$, respectively.
 }
  \label{fig:stiffness}
\end{figure}

Additionally, a larger value of off-diagonal elements indicates that the two categories share more common properties. As a result, a correct judgment of one category is useful for another. For example, \texttt{Percent} category and \texttt{Ordinal} category shared  a common property of ``digit''. We name this relationship between them as \textit{Sibling Categories}, shown in Fig.~\ref{fig:category}.

However, if the off-diagonal elements are negative, it suggests that a gradient step taken with respect to one test entity would increase loss on another test entity with different categories, which we define as  \textit{Overlapping Categories}, shown in Fig.~\ref{fig:category}. This phenomenon usually occurs when two categories have some overlapped entities. For instance, ``\texttt{New York University}'' is usually a \texttt{Location} name, but when ``\texttt{New York University}'' represents as the New York University football team, ``\texttt{New York University}'' is an \texttt{Organization} name.

Particularly, if the off-diagonal elements are close to zero, it means the features of two categories tend to be orthogonal: they share few entities or common properties. We name the relationship of these categories as \textit{Orthogonal Categories}, shown in Fig.~\ref{fig:category}. 

\renewcommand\tabcolsep{1pt}
\begin{table}[!ht]
  \centering \scriptsize
    \begin{tabular}{lcccc|cccc|cccc}
    \toprule
     & \multicolumn{4}{c}{\textbf{CoNLL}} & \multicolumn{4}{c}{\textbf{WNUT}} & \multicolumn{4}{c}{\textbf{ON-BN}} \\
     \cmidrule(lr){2-5} \cmidrule(lr){6-9} \cmidrule{10-13}
    \textbf{Models} & \textbf{F1} & \rotatebox{90}{\textbf{org-loc}} & \rotatebox{90}{\textbf{loc-org}} & \rotatebox{90}{\textbf{misc-o}} & \textbf{F1} & \rotatebox{90}{\textbf{fac.-loc}} & \rotatebox{90}{\textbf{spo.-loc}} & \rotatebox{90}{\textbf{mus.-per}} & \textbf{F1} & \rotatebox{90}{\textbf{ord.-date}} & \rotatebox{90}{\textbf{per-org}} & \rotatebox{90}{\textbf{gpe-loc}} \\

    \midrule
    CcnnWglovecnnMlp& 87.6 & 43.2 & 57.3 & 37.5  & 31.0 & 23.3 & 30.4 & 41.1 & 83.5  & 44.5 & 4.80  & 40.5 \\
    CcnnWglovelstmMlp& 88.4 & 43.3 & 58.9 & 32.9 & 37.8 & 31.1 & 34.6 & 43.8 & 84.1 & 37.5  & 9.09  & 29.7 \\
    CcnnWglovecnnCrf & 89.1 & 50.6  & 64.2 & 29.2 & 34.2 & 34.0 & 34.4 & 37.8 & 85.2 & 27.3 & 30.0 & 51.2 \\
    CnoneWglovelstmCrf & 89.1  & 33.6 & 37.5  & 59.1 & 37.3 & 32.2  & 26.4 & 37.3 & 86.2 & 44.4 & 11.1 & 40.5 \\
    CcnnWglovelstmCrf & 90.3 & 39.9 & 55.1 & 33.7  & 39.2 & 53.7 & 26.9 & 40.8 & 88.6  & 25.0    & 47.2 & 26.5 \\
    CelmWnonelstmCrf & 91.8 & 41.0 & 56.4 & 40.5 & 43.7  & 60.0 & 51.6 & 43.5 & 89.3 & 50.0    & 50.0    & 40.7 \\
    CelmWglovelstmCrf & 92.3 & 48.8 & 54.4 & 35.1 & 44.0 & 51.0 & 43.4 & 46.8 & 90.0 & 28.6 & 33.3 & 33.3\\
    \bottomrule
    \end{tabular}
\caption{Error ratios for hard cases (typical error types) with various NER systems. The detail model architecture is shown in Table \ref{tab:allmodels}.}
  \label{tab:error}
\end{table}

\paragraph{Exp-VI: Exploring the Errors of Hard Cases}  \mbox{} \\ 

As shown in Fig.~\ref{fig:cs-error},
the two sub-figure (a-b) illustrate the consistency and error matrices of the NER model trained on \texttt{CoNLL}.
In the error matrix, the off-diagonal elements of  $Er_{p,q}$ ($p\neq q$) is computed as the number of entity belonging to category $p$ predicted as category $q$, divided by the total number of prediction errors of the category  $p$. The on-diagonal elements of $Er_{p,q}$ ($p=q$) is the accuracy of the category $p$.
Notably, we find that the \textit{consistency} values correlate closely with error ratios based on the Pearson coefficient in  Fig.~\ref{fig:cs-error}-(c).
Taking the marked positions in sub-figures (a-b) for example,
We find that \textbf{if two categories have low \textit{consistency}, the model tends to have difficulty  distinguishing them, and it is easy to mis-predict each other}.
This observation demonstrates that relationships between entity categories influence model's generalization ability.
We additionally find the prospects for further gains from architecture design and knowledge pre-training seem quite limited based Tab.\ref{tab:error}.
To address these issues, more contextual knowledge or prior linguistic knowledge is needed.

\section{ Acknowledgments}
Thanks Jie Fu for useful comments, and thank the anonymous reviewers for their helpful comments. 
This work was partially funded by China National Key R\&D Program (No. 2018YFB1005104, 2018YFC0831105), National Natural Science Foundation of China (No. 61976056, 61532011, 61751201), Science and Technology Commission of Shanghai Municipality Grant  (No.18DZ1201000, 16JC1420401, 17JC1420200), Shanghai Municipal Science and Technology Major Project (No.2018SHZDZX01), and ZJ Lab.

\bibliography{aaai20}
\bibliographystyle{aaai}
\end{document}